# Mining Social Data to Extract Intellectual Knowledge

**Muhammad Mahbubur Rahman**
Department of Computer Science, American International University-Bangladesh
mahbubr@aiub.edu

*Abstract*— Social data mining is an interesting phenomenon which colligates different sources of social data to extract information. This information can be used in relationship prediction, decision making, pattern recognition, social mapping, responsibility distribution and many other applications. This paper presents a systematical data mining architecture to mine intellectual knowledge from social data. In this research, we use social networking site facebook as primary data source. We collect different attributes such as about me, comments, wall post and age from facebook as raw data and use advanced data mining approaches to excavate intellectual knowledge. We also analyze our mined knowledge with comparison for possible usages like as human behavior prediction, pattern recognition, job responsibility distribution, decision making and product promoting.

*Index Terms*— Social Computing, Data Mining, Facebook, Intellectual Knowledge

## I. Introduction

With the advance of Internet and Web technologies, the accessibility of social data through social networks, blogs, forums and news sites has increased rapidly [1-3] and has pulled much attention all over the world in recent research [4]. These social data can be used in marketing, decision making, destabilizing terrorist networks, behavior evolution, discovering social relationship from multiple entities [5, 7, 8, 11, 15] as well as many other applications [6]. But those social data need to be reflected for suitable usages. To process social data for suitable usages, social network analysis is one of the most important issues. Many data mining algorithms can be used to analyze social networks [9-10]. For last few years, some research works [12-14] have been developed on social networks. But all of those researches are mainly for social interaction and connectivity analysis. Those researches have not been focused on mining intellectual knowledge and comparison analysis. Again, none of them has been done on the analysis of individual social data in the social usages. In this paper, we use different data mining approaches to analyze social networking sites such as facebook and we also apply some mining algorithms to extract intellectual knowledge from mined social data. We developed a systematical architecture for whole procedure from data collection to visual representation of intellectual knowledge. Moreover, we also include a comparative result analysis on our extracted intellectual knowledge which can be used in any social application such as human behavior prediction, marketing, pattern recognition, product promoting and distribution of job responsibilities.

The structure of this paper is organized as bellow. In section II, we discuss different methodologies and algorithms that are used in this research work. Section III describes the architecture of our developed system. In section IV, we represent and compare the result of our experiment in graphical form. Section IV also includes comparative analysis of our intellectual knowledge for social usages. Section V is the concluding section of our research paper.

## II. Methodology

In our research, we follow several data mining techniques to extract knowledge from social data. Initially a group of attributes have been selected based on available user information towards friends, groups and public on social networking site (Facebook). The attributes which may imply a user's activities have been considered. A list of attributes is given in Table 1. Then necessary data has been collected from Facebook for those attributes. Facebook API Key and Application Secret key have been generated based on Facebook API to get access on Facebook user database. Collected data has been stored in our secondary data-base. Then text data is converted into discrete value using different data mining techniques and WEKA file has been generated. Different techniques and algorithms that have been used on data to generate WEKA file are described briefly.

### A. Features

Taking whole content from any *text* attribute and filtering the content by removing common words, keywords are generated. We consider each keyword as a term and calculate their frequency in that *text* attribute using *term frequency*.

The *term frequency (TF)* in any *text* is simply the number of occurrences in the *text*. This count is usually normalized to prevent a bias towards longer *text* (which may have a higher *term frequency* regardless of the actual importance of that term in the *text*) to give a measure of the importance of the term $t_i$ within the particular *text* $d_j$.

$$tf_{i,j} = \frac{n_{i,j}}{\sum_k n_{k,j}} \quad (1)$$

Where $tf_{i,j}$ is the *term frequency (TF)* in the *text i* for a particular term *j*, $n_{i,j}$ is the total number of occurrences of the considered term in *text* $d_j$, and the denominator is the total number of occurrences of all terms in *text* $d_j$.

Then a list of terms having higher term frequency (sorted in descending order) has been selected as features for our experiments.

Table 1. Attribute List

| Attribute Name | Data Type |
|---|---|
| Birthday | Date |
| About_me | Text |
| Activities | Text |
| Gender | Text (Discrete) |
| Interests | Text |
| Wall_count | Number |
| Political | Text |
| Music | Number |

### B. Class

We have chosen a list of text documents of different categories as sample classified text. For example, when we consider about me attribute as our text data, we have taken more than 50 about me text of each category as sample about me text where class level of each text is known. For about me attribute, different class levels are Aggressive, Honest, Romantic, Sincere, Dishonest, Friendly, Eager_to_Learn, Conservative, Emotional and Lazy. Each of these class levels defines user's information.

### C. Distance Matrix

A *square matrix* has been generated where number of rows is total number of sample classified known *text* documents and number of columns is total number of selected features from the taken *text* attribute. Each value of every row is calculated by taking square of difference between total numbers of occurrence of each feature in the taken *text* attribute and number of occurrence of that feature in the corresponding sample chosen *text* document. Another matrix called *distance matrix*, $D_m$ is generated based on square matrix using Euclidian Distance. Number of rows of *distance matrix* is total number of sample classified known text documents. Each row contains distance between taken *text* attribute and corresponding sample chosen *text* document.

The Euclidean Distance between taken *text* attribute, $d_j$ and corresponding sample chosen *text* document, $s_i$ is the length of line segment $\overline{d_j s_i}$. Both $d_j$ and $s_i$ have multiple features that means multiple dimensions. So Euclidean Distance is calculated by following formula.

$$D(d_j, s_i) = \sqrt{(d_{j1}-s_{i1})^2 + (d_{j2}-s_{i2})^2 + (d_{j3}-s_{i3})^2 + \ldots + (d_{jn}-s_{in})^2} = \sqrt{\sum_{k=1}^{n}(d_{jk}-s_{ik})^2} \quad (2)$$

Where n is the total number of selected features.
And *distance matrix* is calculated by Equation (3).

$$D_m = \begin{bmatrix} D_1(d_1, s_i) \\ D_2(d_2, s_i) \\ \ldots \\ \ldots \\ D_t(d_n, s_i) \end{bmatrix} \quad (3)$$

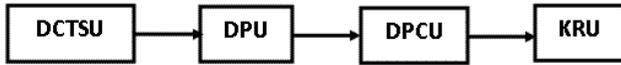

Fig.1 Block Diagram of Different Units

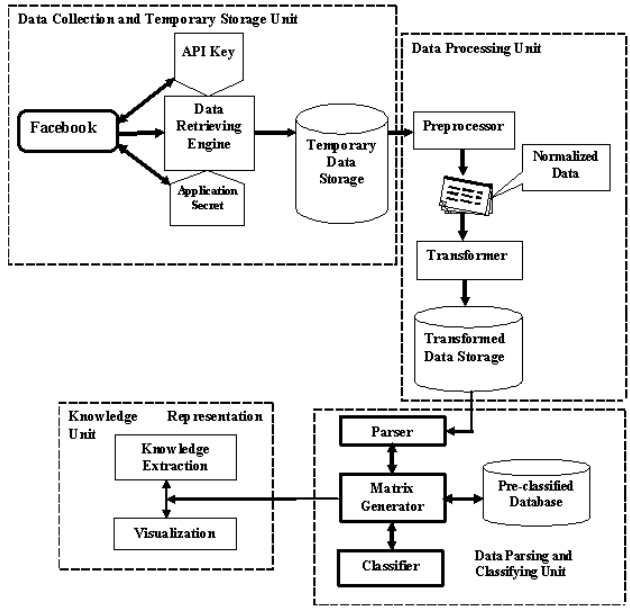

Fig.2 System Architecture

rithms for data analysis and predictive modeling. To analysis data using Weka, input file should be generated in a specific format. Attribute-Relation File Format (ARFF) is one of the supported file format. An ARFF file is generated by our system where different attributes have been chosen to analysis and predict user behavior on social networking site Facebook.

### III. System Architecture

The whole system architecture has several sub units. These are Data Collection and Temporary Storage Unit (DCTSU), Data Processing Unit (DPU), Data Parsing and Classifying Unit (DPCU) and Knowledge Representation Unit (KRU). The block diagram of different units is given in Figure 1. Figure 2 represents the whole system architecture of our developed system. Different units of system architecture are described briefly.

#### A. Data Collection and Temporary Storage Unit collects data from Facebook

It has data retrieving engine and temporary data storage. Data retrieving engine directly connects to Facebook server. With the help of API key and Application Secret key provided from Facebook, it logs in to Facebook accounts. It retrieves user's information like as content of about me, user's wall, age, gender, comments, number of wall count, number of music and so on. It then stores collected data into temporary data storage. Data retrieving engine is developed by PHP and Apache web server. Temporary data storage is designed by MYSQL database server.

#### B. Data Processing Unit

Data Processing Unit converts retrieved data into transformed data which is used by Data Parsing and Classifying Unit. It has preprocessor, transformer and transformed data storage. Preprocessor takes data from temporary data storage and processes that into normalized data. This normalized data is used by transformer to convert that into transformed data. Transformed data is the optimized data which has no error, noisy and missing information. Transformed data is stored in transformed data storage which is also a MYSQL database.

#### C. Data Parsing and Classifying Unit

Where the total numbers of sample classified known *text* documents is $t$ and $s_i$ is the taken *text* attribute.

#### D. Classifier

*Distance matrix* is sorted in ascending order. *K-nearest neighbor (*k-NN*)* classifier is applied on *distance matrix* to classify taken *text* attribute based on the class levels from classified known *text* documents. In pattern recognition, *K-nearest neighbor* is an algorithm for classifying objects based on closest training samples in any feature space. K-NN is a type of instance-based learning or lazy learning which classifies any object by a majority vote of its neighbors. Using k-NN, lowest k distance values are selected with their class levels and taken *text* attribute is classified based on majority class levels.

#### E. WEKA file Generation

Weka (Waikato Environment for Knowledge Analysis) is a package of machine learning algorithms that contains a collection of graphical applications and algo-Data Parsing and Classifying Unit is the most important unit of the whole system. It has parser, matrix generator, classifier and Pre-classified sample Database of known class levels. Parser collects data from transformed data storage of data Processing Unit. Initially it processes text data like as about me attribute's value of any user. At the starting of parsing it removes all common words from text. It has a common words list. It generates keyword, term and term frequency using (1). Finally it selects features based on term frequency. Selected features are used for the experiment. Matrix generator uses selected features and pre-classified sample database to generate distance matrix using (3). It uses (2) to calculate distance between any feature and pre-classified text documents. Classifier uses distance matrix to classify text attributes. K-nearest neighbor (k-NN) algorithm is used by classifier to classify text attributes. Classifier also uses other numeric attributes and range value attributes such as age, wall count, music count, activity and interest count and so on to classify them into specific class levels.

### D. Knowledge Representation Unit

It has data retrieving engine and temporary data storage. Data retrieving engine directly connects to Facebook server. With the help of API key and Application Secret key provided from Facebook, it logs in to Facebook accounts. It retrieves user's information like as content of about me, user's wall, age, gender, comments, number of wall count, number of music and so on. It then stores collected data into temporary data storage. Data retrieving engine is developed by PHP and Apache web server. Temporary data storage is designed by MYSQL database server.

### IV. Knowledge Extraction and Discussion

Our system collects data of 1753 users of different ages and genders. It filters out erroneous and missing data. After filtering, 1340 users' data have been used for the experiment. In this section, knowledge extraction and comparison of different attributes such as about me, age, gender, wall count, number of music, number of activities and interests are presented. These knowledge and comparison can be used for decision making, classifying human being, human behavior.

### A. On "about me" with respect to various age ranges

Our system analyzes content of **"about me"** as one of the attributes. It uses Eq. (1), (2) and (3) to convert content of **"about me"** into some discrete numerical values. Then it classifies those data based on our pre-defined class levels. Figure 3 represents classification of users up to age level 19 years. It implies that the height numbers of Facebook users of this age group are honest, the second height is romantic and the third height is conservative. Figure 4 depicts pie chart representation of users of age range 20 to 32 years. It entails that the height numbers of Facebook users of this age group are aggressive, the second height is honest and the third height is romantic. Figure 5 is a pie chart representation of age range 33 to 45 years. It implicates that the height numbers of Facebook users of this age group are sincere, the second height is conservative and the third height is aggressive. Figure 6 represents classification of age over 45 years. It implies that the height numbers of Facebook users of this age group are sincere, the second height is honest and the third height is conservative. Figure 7 visualizes users who have hidden their age to others. It implies that the height numbers of Facebook users of this age group are conservative, the second height is dishonest and the third height is aggressive. This intellectual knowledge about any person can be used to predict human behavior, nature and personality which are important to get idea about any human being. This knowledge can also be used to define our social model in different stages.

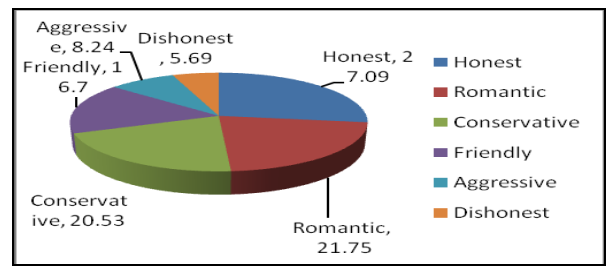

Fig.3 Age Range up to 19 years

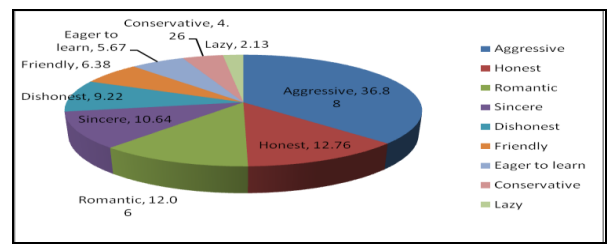

Fig.4 Age Range 20 - 32 years

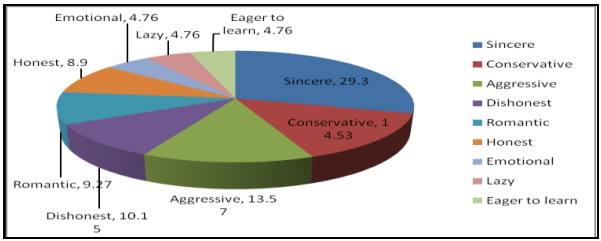

Fig.5 Age Range 33 - 45 years

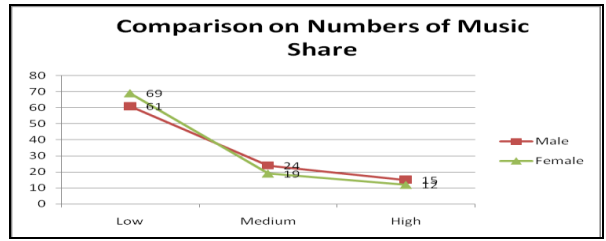

Fig.10 Comparison on Numbers of Music Share

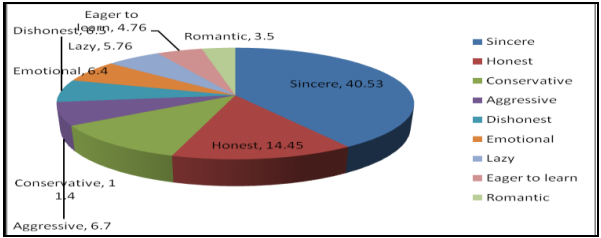

Fig.6 Age over 45 years

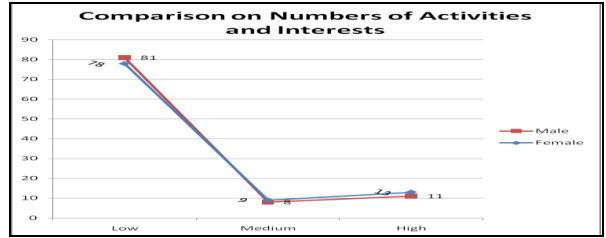

Fig.11 Comparison on Numbers of Activities and Interests

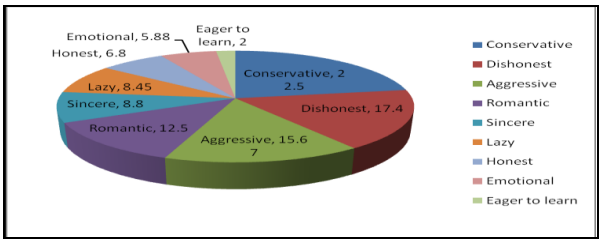

Fig.7 Age is hidden to others

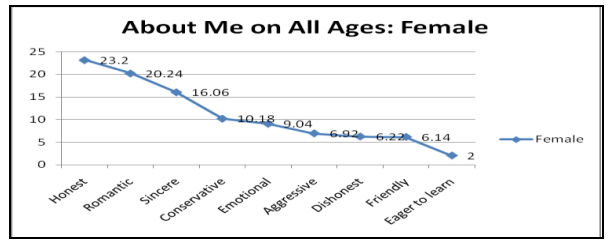

Fig.12 Female of all Ages

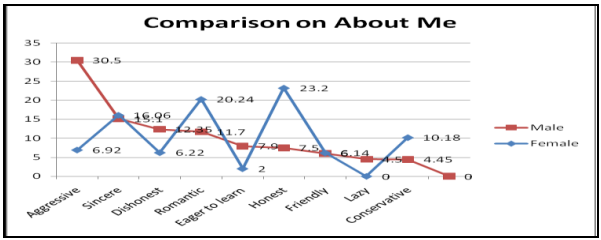

Fig.8 Comparison on About Me for Male and Female

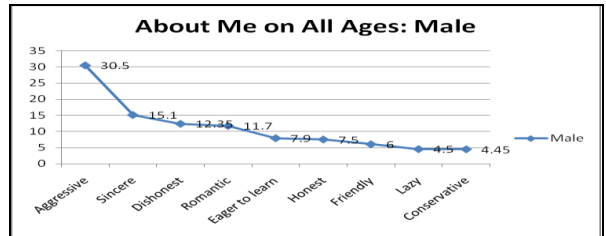

Fig.13 Male of all Ages

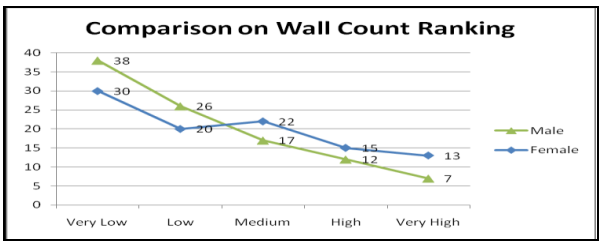

Fig.9 Comparison on Wall Count Ranking

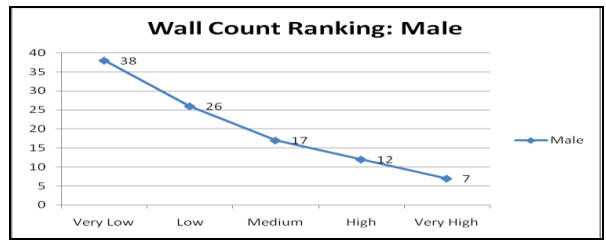

Fig.14 Wall Count Ranking for Male

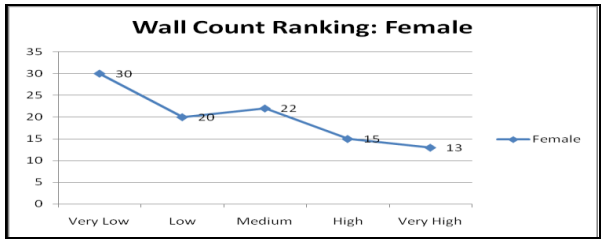

Fig.15 Wall Count Ranking for Female

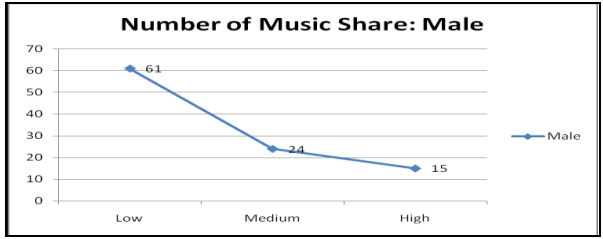

Fig.16 Number of music share for Male

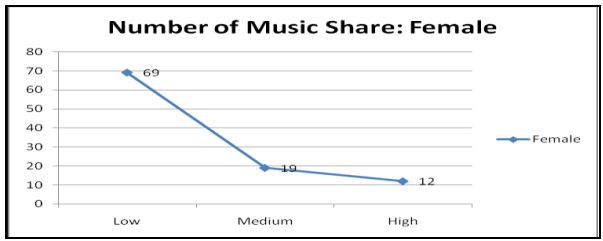

Fig.17 Number of music share for Female

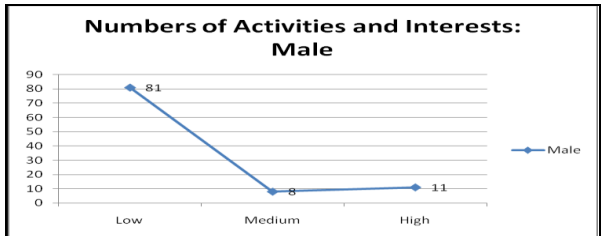

Fig.18 Numbers of Activities and Interests for Male

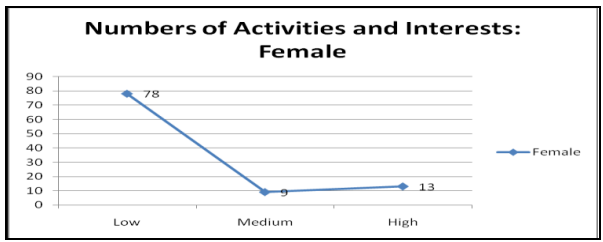

Fig.19 Numbers of Activities and Interests for Female

Table 2. Wall Count Ranking Class

| Range | Group Class Name |
|---|---|
| <10 | Very Low |
| 10 to 50 | Low |
| 51 to 100 | Medium |
| 101 to 200 | High |
| >200 | Very High |

Table 3. Music Share Class

| Range | Group Class Name |
|---|---|
| <5 | Low |
| 6 to 15 | Medium |
| >15 | High |

Table 4. Activities and Interests Class

| Range | Group Class Name |
|---|---|
| <5 | Low |
| 6 to 15 | Medium |
| >15 | High |

### B. On "about me" with respect to male and female of all ages

Our system analyzes all users' data and divides into two major groups according to gender. It also analyzes "**about me**" of all male and female separately and compares. Figure 12 and Figure 13 represent line charts of all ages on about me attribute for female and male facebook users respectively. Figure 8 depicts comparison line chart of male and female users of all ages on about me attribute. It implies that the height numbers of male Facebook users are aggressive, the second height is sincere and the third height is dishonest where as the height numbers of female Facebook users are honest, the second height is romantic and the third height is sincere. This analytical knowledge can be used in behavior prediction of male and female in the society. It also can be used in any organization to distribute job responsibilities among employees of different genders which helps the organization to get the maximum pay off from its employees.

### C. On "wall count" with respect to male and female

Our system collects total number of post on the wall of all users as well as contents of each post. It defines several groups based on continuous number of post. The groups' names are Very Low, Low, Medium, High and Very High. The group classification is given in the Table 2. Figure 14 and Figure 15 represent line charts of male and female facebook users respectively on wall count. Figure 9 depicts line chart of male and female comparison on all wall counts. Content of each post is also analyzed to get correlation between posts and users. This intellectual knowledge can be used to discover human thinking capability in any specific topic.

### D. On "music share" with respect to male and female

Our system collects total number of sharing music of all users as well as name of music. It defines several groups based on number of sharing music. The groups' names are Low, Medium and High. The group classification is given in the Table 3. Figure 16 and Figure 17 represent line charts of male and female facebook users respectively on music share. Figure 10 sketches line chart of male and female comparison on number of music share. This comparative knowledge can be used to extract information about any famous music and interests of users on music based on genders.

### E. On "Activities and Interests" with respect to male and female

Our system collects total numbers of activities and interests of all users as well as content of each activity and interest. It defines several groups based on numbers of activities and interests. The groups' names are Low, Medium and High. The group classification is given in the Table 4. Figure 18 and Figure 19 represent line charts of male and female facebook users respectively on activities and interests counts. Figure 11 limns line chart of male and female comparison on all activities and interests counts. This extracted knowledge can be used for human behavior and pattern recognition with respect to male and female in the society.

## V. Conclusion

Social data mining is an interesting and challenging research to mine intellectual knowledge which can be used in human behavior prediction, decision making, pattern recognition, social mapping, job responsibility distribution and product promoting. In this paper, we present a systematical data mining architecture to collect social data from facebook, normalize collected data, transform normalized data, parse transformed data, classify parsed data, extract intellectual knowledge from classified data, compare and visualize the knowledge for appropriate usages.

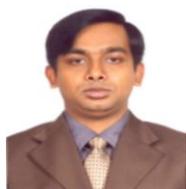

**Muhammad Mahbubur Rahman** is currently working as an Assistant Professor in the department of Computer Science at American International University-Bangladesh. He received his master degree in Information Technology from University of Dhaka, Bangladesh. His research interests include Data Mining, Machine Learning, Bioinformatics, Game Theory and Artificial Intelligence. Currently his active research works are in bioinformatics and data mining. He has several years of working experience in local and international telecommunication and software companies.